\def\year{2020}\relax
\documentclass[letterpaper]{article} 
\usepackage{aaai20}  
\usepackage{times}  
\usepackage{helvet} 
\usepackage{courier}  
\usepackage[hyphens]{url}  
\usepackage{graphicx} 
\usepackage{graphicx}  
\usepackage{multirow}
\usepackage{amsmath,bm}
\usepackage{amsfonts}
\usepackage[farskip=0pt]{subfig}
\usepackage{amssymb}
\usepackage[ruled,vlined]{algorithm2e}
\usepackage{booktabs}
\title{Potential Passenger Flow Prediction: A Novel Study for Urban Transportation Development}

\author{Yongshun Gong\textsuperscript{\rm 1}, Zhibin Li\textsuperscript{\rm 1}, Jian Zhang\textsuperscript{\rm 1}, Wei Liu\textsuperscript{\rm 1}, Jinfeng Yi\textsuperscript{\rm 2}\\ 
	\textsuperscript{\rm 1}Faculty of Engineering and IT, University of Technology Sydney, Sydney, Australia\\
	\textsuperscript{\rm 2}JD AI Research, Beijing, China\\
	\{yongshun.gong, zhibin.li\}@student.uts.edu.au; \{jian.zhang, wei.liu\}@uts.edu.au; yijinfeng@jd.com 
}
 
\begin{document}

\maketitle

\begin{abstract}
Recently, practical applications for passenger flow prediction have brought many benefits to urban transportation development. With the development of urbanization, a real-world demand from transportation managers is to construct a new metro station in one city area that never planned before. Authorities are interested in the picture of the future volume of commuters before constructing a new station, and estimate how would it affect other areas. In this paper, this specific problem is termed as potential passenger flow (PPF) prediction, which is a novel and important study connected with urban computing and intelligent transportation systems. For example, an accurate PPF predictor can provide invaluable knowledge to designers, such as the advice of station scales and influences on other areas, etc. To address this problem, we propose a multi-view localized correlation learning method. The core idea of our strategy is to learn the passenger flow correlations between the target areas and their localized areas with adaptive-weight. To improve the prediction accuracy, other domain knowledge is involved via a multi-view learning process. We conduct intensive experiments to evaluate the effectiveness of our method with real-world official transportation datasets. The results demonstrate that our method can achieve excellent performance compared with other available baselines. Besides, our method can provide an effective solution to the cold-start problem in the recommender system as well, which proved by its outperformed experimental results.
\end{abstract}

\section{Introduction}
With the growth of intelligent transportation systems, passenger flow prediction models concentrate on discovering the volume of crowds and mobility patterns that best serve people's daily life \cite{pan2019urban,zhang2019flow}. Recent advances in passenger flow prediction are focusing mainly on next time interval flow conditions with time evolves \cite{gong2018network,sun2015integrated}. If a brand-new metro station is inserted into the original metro network, existing predictors have to collect a large amount of latest transactional data to ensure normal operation. However, a real-world requirement from transportation authorities is that they want to obtain the potential passenger flows (PPF) of a planned city area in advance (i.e., before constructing a station in this area). It is significant for the urban traffic development and transportation management, as it can provide insights for the site selection of stations and analysis of passenger movement patterns, as well as give the potential crowd warning.

\begin{figure}[t]
	\centering
	\includegraphics[width=1\linewidth]{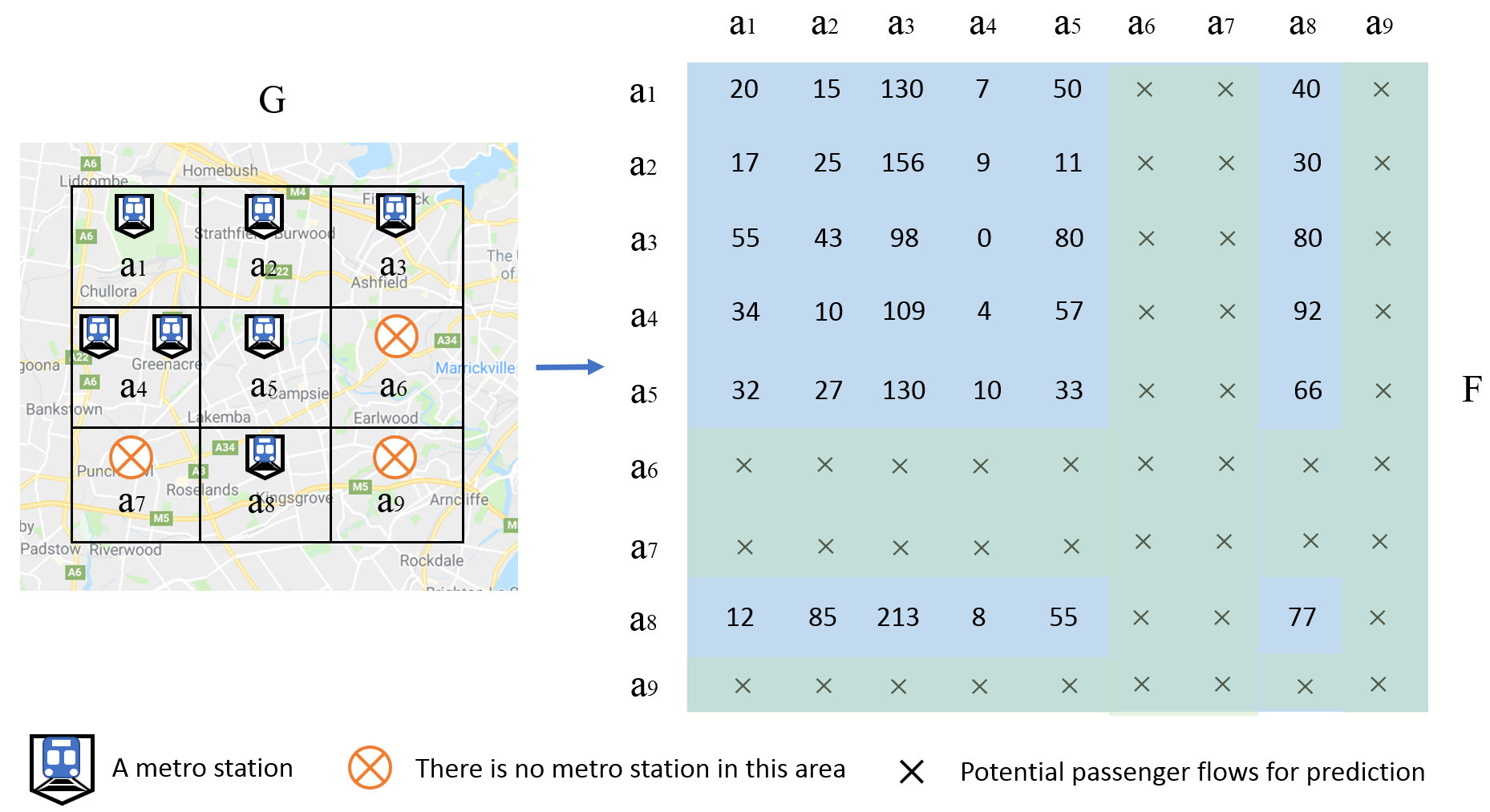}
	\caption{The example of PPF prediction problem. We aim to forecast the passenger flows of target areas (e.g., $a_6$, $a_7$, $a_9$) across the entire city network.}	
	\label{figure_PPF_example}
\end{figure}

In the PPF prediction task, concentrating solely on the entrance and exit potential flows does not provide adequate information, authorities also desperately want to master the distribution of predicted PPF, i.e., forecast the number of potential passengers moving to different destinations. It is utmost important to find how will the new station affect other areas. For instance, Figure \ref{figure_PPF_example} illustrates an example of the PPF prediction problem. A city region is partitioned into nine areas\footnote{We use grids for clear and simple illustration, the real partition standard is explained in the section of data description.}, six of them have metro stations (termed as known areas), and three have not constructed yet (termed as target areas). The right part of Figure \ref{figure_PPF_example} presents an origin-destination (OD) matrix (each row point is the origin area and column points are destinations), e.g., $F$($a_1$, $a_3$) = 130 indicates that there are 130 passengers departure from $a_1$ and are going to the $a_3$. PPF task aims to make an accurate prediction for the target areas in one period (e.g., rush hours) that completes the crowd flows between them and known areas.

To date, limited studies considered the OD passenger flow prediction problem \cite{gong2018network,wang2019origin}, and to the best of our knowledge, none of existing techniques can forecast PPF across the entire city. It is a novel problem
and a real urban developing demand that faces several major challenges: (1) Considering the number of passenger flows and their final destinations simultaneously. (2) Analogously to the cold-start problem in the recommender system \cite{lam2008addressing}, it is hard to infer the preference of a new user from the known data. In our problem, a new station in the target area can be similarly regarded as a new user. (3) Since the PPF is a spatial-temporal mining problem, spatial and temporal information should be taken into account appropriately. 

To resolve this novel and significant problem, in this paper, we devise a multi-view localized correlation learning model for the PPF prediction (MLC-PPF for short). To leverage the spatial information, we first construct a localized similarity matrix which associates with the real geographical neighbors and regional properties (e.g., business or residential regions). The intuition behind this strategy is from the First Law of Geography \cite{tobler1970computer}, i.e. \textit{``Everything is related to everything else, but near things are more related than distant things''}. Second, a novel weighted correlation learning strategy is proposed. At last, to improve the prediction accuracy and well handle the cold-start challenge, we draw the side information from urban statistical data, where each area has a multi-view features to guide the learning process. In summary, our main contributions are shown as follows:

\begin{itemize}
\item We formulate the PPF prediction problem and provide the first attempt on forecasting passenger flows for urban transportation development.

\item We propose a multi-view localized correlation learning method to provide a solution for the PPF prediction that can learn localized correlations via a multi-view learning process.

\item We show that our method can be transferred to the classic cold-start problem in the recommender system. It achieves a superior result that gives a new perspective for relevant tasks.

\item We conduct extensive prediction experiments on a large real-world transactional dataset and show that our model outperforms other available algorithms.
\end{itemize}

\section{Related Work}

\subsection{Passenger Flow Prediction}

Most existing passenger flow prediction models focused on forecasting entrance/exit flows at certain stations or areas, neglecting the crowd flows across different stations. \cite{chen2011exploring,wei2012forecasting,ni2017forecasting}. Wei et al. \cite{chen2011exploring} developed an effective short-term passenger flow prediction model to explore the time variants and capture dynamic patterns on a single subway line. Subsequently, a modified approach is proposed based on the neural network, which aims to solve the same entrance/exit crowd flow prediction task in a few metro lines \cite{wei2012forecasting}. Ni et al. \cite{ni2017forecasting} used auxiliary information, such as social media events, to improve the forecast performance.

One of the research hotspots is named the city network-wide crowd flow prediction, which is a significant task for the modern transportation management \cite{DengSDZYL16}. Nowadays, some of methods were focusing on forecasting the citywide crowd flows. \cite{ma2016mobility} devised a series of visualization approaches to show the flows' dynamic changes in the networks. Zhang et al. proposed the deep learning models based on the ResNet to predict crowd inflows and outflows of the entire city regions. \cite{zhang2018predicting,zhang2019flow}. The Probabilistic model is an effective approach to estimate the traffic speed. For example, \cite{zhan2016citywide} and \cite{lin2017road} used trajectory data to estimate citywide traffic volume via probabilistic graphical models. \cite{gong2018network} proposed an effective method based on online latent space learning to predict the crowd flow distribution, i.e., forecast the OD pairs and the quantity of passenger flows simultaneously. To the best of our knowledge, none of existing crowd flow prediction methods considered the PPF problem studied in this paper.

Other relevant studies, such as \cite{Hsun2015Inferring}, are point-based prediction model, not in a matrix formulation. \cite{Hsun2015Inferring} selects \textit{k} points to predict \textit{k} values. But in our task, \textit{k} target areas are required \textit{nk} prediction values, where \textit{n} is the number of known areas. It is because we also need to consider the crowd flows between each area.

\begin{figure*}[t]
	\centering
	\includegraphics[width=1\linewidth]{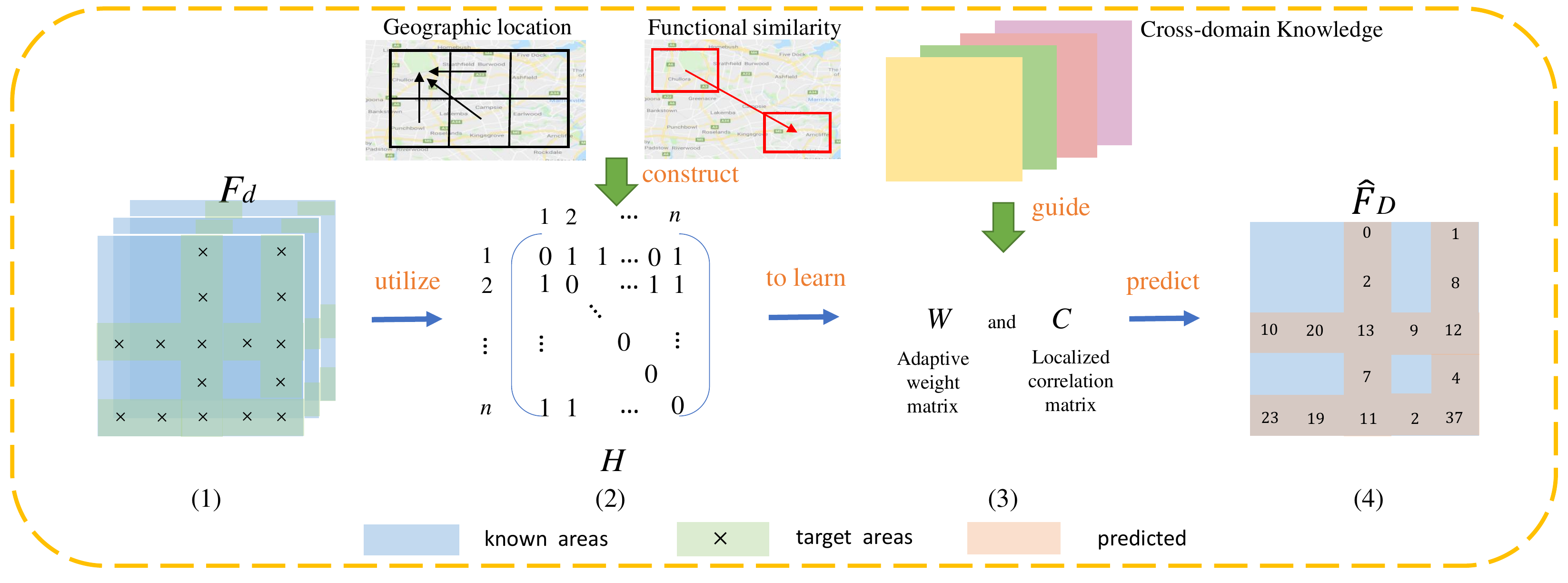}
	\caption{The flowchart of our proposed model. In the learning process, given a set of previous PPF matrices \{$F_d$\}, MLC-PPF learns the localized correlation matrix $C$ and adaptive-weight $W$ via a $k$-nearest indicator matrix $H$. The cross-domain knowledge is utilized to guide the
	updating of $C$. Then, the target prediction can be inferred by Algorithm 1. }	
	\label{figure_flowchart}
\end{figure*}

\subsection{Multi-view Learning}

Traditional passenger flow mining usually deals with data from a single view. Recently, there exists a diversity of datasets from different sources in various domains with multiple views \cite{zheng2015methodologies,li2019sample}. The multi-view learning algorithm is widely recognized as an effective way of solving the cross-domain problem, that features from different views can be served for the target domain learning process \cite{singh2008relational,xu2013survey,elkahky2015multi}. \cite{xu2015multi} proposed a matrix co-factorization based method (MVL-IV) to embed different views into a shared subspace, such that the incomplete views can be estimated by the information on observed views. To connect multiple views, MVL-IV assumed that different views have distinct ‘feature’ matrices (i.e., $\{\bm H_{i}\}^m_{i=1}$), but correspond to the same coefficient matrix (i.e., $\bm W$). The tensor-based methods, such as \cite{hu2013personalized}, \cite{hu2016learning} \cite{taneja2018cross} were proposed to address the cross-domain recommendation problem. They devised a cross-domain triadic factorization model to learn the triadic factors for user, item and domain, where the item dimensionality varies with domains. The above approaches and other similar methods \cite{rendle2009learning,xiong2010temporal} cannot address our PPF prediction problem directly because they are not formulated for the passenger flow prediction task. However, since they can handle the cold-start problem by utilizing the cross-domain knowledge, an illuminative clue is educed.

In conclusion, none of relevant studies can solve the PPF prediction problem directly. Accordingly, this paper aims to design a reliable approach for PPF prediction with cross-domain knowledge involved.

\section{Problem Statement}

Focusing on the PPF prediction problem, every origin-destination among areas needs to be recorded. We formulate the OD passenger flow network as a fully connected graph $G$ = ($A$, $E$), where $A$ is a set of vertexes and $E$ is the set of edges. $a_{i}\subseteq V$ records the $i$-th origin or destination area, and an edge $e(v_{i}, v_{j})$ denotes an origin-destination flow from area $a_{i}$ to $a_{j}$. The value of each edge $e(a_{i}, a_{j})$ is associated with the observed flow $f(a_{i}, a_{j})$, i.e., $f(a_{i}, a_{j})$ is the total number of passengers that departure from $i^{th}$ area and are going to the $j^{th}$ area. Then, \textit{G} can be represented by PPF matrix $F = (f_{ij})$, where $f_{ij}$ = $f(a_{i}, a_{j})$. The example of $G$ and $F$ is shown in Figure \ref{figure_PPF_example}. $f_{31}$ = 55 means that 55 passengers leave from $a_1$ and theirs' destination is $a_2$.

In the real-world scenario, one area may have several stations. In this case, we calculate the passenger flows of these stations together to present the total flows of the area. We consider three specific and useful time periods to predict PPF, which will help the authorities to do a better temporal analysis of transportation development. The three periods are morning rush hour (7.00 AM - 9.00 AM), afternoon rush hour (5.00 PM - 7.00 PM), and non-rush hour (2.00 PM - 4.00 PM).  

Furthermore, traffic periodicity is a very important factor for relevant studies. Crowd flows also
represent the stable and daily periodic properties, especially on weekdays. To extract the temporal information and make a more general prediction, we consider a series of previous daily PPF matrices ($F_1, F_2, \cdots, F_D$) in the same time period to predict the PPF matrix of target areas ($\hat{F_D}$) for the day $D$. Note that, the prediction is not limited in the $D$-th day, the target can be changed easily based on the real requirement.

To best simulate the crowd flow changes when picking the target areas, in this paper, we tracked all trajectories of passengers, from origins to their destinations. For example, if area $a_1$ is selected as a target area, all the departure crowd flows from $a_1$ will add to its closest area (e.g., $a_2$) to best simulate the people’s choice. In this way, the PPF is learned by the crowd flows under the assumption that the original passengers from $a_1$ will departure from its closest neighbor $a_2$.

\section{The Proposed Method}

In this section, we propose our PPF prediction model MCL-PPF. We will describe the strategy of localized correlation learning with adaptive-weight, and how to leverage the cross-domain multi-view information to improve our work. Figure 2 shows the flowchart of our model.

\subsection{Localized Correlation Learning}

PPF prediction problem is a spatially related task that the more similar between two areas, the more correlations of passenger flow condition they have. Assuming that a city is partitioned into $n$ areas, including known and target areas. $F\in \mathbb{R}^{n\times n}$ presents the PPF matrix, and $F^k\in \mathbb{R}^{n\times n}$ presents the localized flow matrix of $F$, where the $i$-th row of $F^k$ is a combination of its $k$-nearest area passenger flows. In that way, we formulate the function to learn the localized correlation, which can be expressed as:

\begin{equation} \label{equ-basic}
\begin{split}
&\min _{C} ~~\frac{1}{2} \sum_{d=1}^D \left \| F_d - F_d^kC \right \|_F^2  \\ \quad  &{\rm s.t.} ~~P_{\Omega}(F_d) = P_{\Omega}(F_d^k)
\end{split}
\end{equation}
where $C\in \mathbb{R}^{n\times n}$ is the localized correlation matrix that learns the transformation from $F^k_d$ to $F_d$ in each period of day $d$, $D$ is the total number of days; $P_{\Omega}(\cdot)$ stands for the projection of all observed passenger flows from the known area set $\Omega$; $\left \| \cdot  \right \|_F$ is Frobenius norm of matrix.

Now, we will discuss how to build $F^k_d$ of one period in a day. The physical distances among areas need to be considered first. Moreover, the development of a city gradually fosters different functional regions, such as business and educational areas, where the areas belonging to the same functional region will have strong connections with their properties \cite{zheng2014urban}. 

Thus, the similarities among areas should take into account the above two standards. To this end, we build two distance metrics from the real geographic location and regional similarity. The distance metric between $i$-th and $j$-th areas is shown as follows:

\begin{equation} 
\label{equ-similarity}
s_{i,j} = 2 - (\frac{dist_{i,j}}{max(dist_{i,:})} + \frac{dist_{i,j}^{'}}{max(dist_{i,:}^{'})})
\end{equation}
where $dist_{i,j}$ is the geographic distance between $i$-th and $j$-th areas; and $dist_{i,j}^{'}$ presents the Euclidean distance which is calculated by intrinsic features of areas (e.g., point of interest attributes).

After having gotten the $s_{i,j}$, the $k$ neighbors of the $i$-th area can be picked. Then, we construct an indicator matrix $H$ for the $k$-nearest neighbors of all areas where each row indicates the position of its $k$-nearest known areas. For example, in the stage 2 of Figure 2, the first row of $H$ illustrates that $a_2$, $a_3$, and $a_n$ are the $k$-nearest areas of $a_1$ if $k$ is setting to 3. Accordingly, for each day $d$, the localized flow matrix $F^k_d$ can be represented as $F^k_d = (H\odot W)F_d$, where  $W$ is an adaptive-weight matrix that learns the different weights of $k$-nearest areas. To this stage, the localized correlation learning process is shown as Eq. (\ref{equ-k-nearest}).

\begin{equation} \label{equ-k-nearest}
\begin{split}
\min _{C,W} ~~\frac{1}{2} &\sum_{d=1}^D \left \| F_d - (H\odot W)F_dC \right \|_F^2 \\  \quad  &{\rm s.t.}~~P_{\Omega}(F_d) = P_{\Omega}(F_d^k)
\end{split}
\end{equation}
where $\odot$ is the entrywise product; The loss function aims to learn the localized correlations matrix $C$ and weight matrix $W$ simultaneously.

\subsection{Improvement by Cross-domain Learning Process}

As mentioned above, there are various functional regions of a city. Thanks to the urban statistical data, the passenger flow similarities among different areas can be reviewed from this cross-domain perspective. Based on the phenomenon that the similar functional regions have the similar passenger flows (e.g., the business regions have a large number of entrance flows during the morning rush hour, and people leave from residential areas in the same time span), we leverage such information to guide the localized correlation learning process.

The statistical data have multiple views to record the differences between areas. For example, the economy view reveals the economic features, such as the number of industries and employee statistics; and the population view consists of detailed population information. Let \{$X_1, X_2, \cdots, X_V$\} denote the multi-fold views of statistical data, where $X_v \in \mathbb{R}^{n\times m_v}$, the row of $X_v$ denotes the area and column denotes the feature. To improve the prediction performance, cross-domain knowledge is involved as guidance, which can be formulated as:

\begin{equation} \label{equ-improved}
\begin{split}
\min _{C,W} ~~\frac{1}{2} &\sum_{d=1}^D \left \| F_d - (H\odot W)F_dC \right \|_F^2 + \frac{\lambda}{2} \sum _{v=1}^V \left \| X_v - CX^k_v \right \|_F^2  \\ &\qquad\qquad {\rm s.t.}~~P_{\Omega}(F_d) = P_{\Omega}(F_d^k)
\end{split}
\end{equation}
where $\lambda$ is the regularization parameter; $X_v^k = HX_v$ denotes the localized matrix of $X_v$.

After solving Eq. (\ref{equ-improved}), the learned matrices $W$
and $C$ can be used to make the prediction. The predicted PPF of target areas in $D$-th day  is:

\begin{equation} \label{equ-predict}
\hat{F_D} = (\bm 1-Y)\odot ((H\odot W)F_DC)
\end{equation}
where $Y$ is an indicator matrix whose entry ($i,j$) is one if $ F(i,j)$ is observed and zero otherwise.

To this stage, the OD passenger flows in the target areas are learned by the above processes, i.e., predict each row of target areas. However, the column of target areas revealing how much crowds arrived at these areas needs to be predicted with
a slight modification. That is, replace $F_d$ with $F_d^{\top}$ in Eq. (\ref{equ-improved}) to learn the localized correlation from the other side. It can be solved in a likewise manner. Thus we only presented the optimization strategy of Eq. (\ref{equ-improved}) due to the page limitation.

\subsection{Learning and Prediction}

Eq. (\ref{equ-improved}) is a complex non-convex problem. But the loss function associated with Eq. (\ref{equ-improved}) is convex regarding $C$ with fixed $W$ and vice verse. We can optimize them alternatively until convergence (e.g., alternating least squares (ALS)). A straightforward way to minimize the loss function is through the gradient method.

Considering $C$ while $W$ is fixed, Eq. (\ref{equ-improved}) can be rewritten as follows:

\begin{equation} \label{equ-gC-1}
\begin{split}
\mathcal{L}  = &\frac{1}{2} \sum_{d=1}^D Tr((F_d - (H\odot W)F_dC)(F_d - (H\odot W)F_dC)^{\top}) \\ & + \frac{\lambda}{2} \sum_{v=1}^V Tr((X_v - CX^k_v)(X_v - CX^k_v)^{\top})
\end{split}
\end{equation}

Taking the derivative of $\mathcal{L}$ with respect to $C$, we can get
gradient $gC$:

\begin{equation} \label{equ-gC}
\begin{split}
gC = &\sum_{d=1}^D ((H\odot W)F_d)^\top((H\odot W)F_dC- F_d) \\ &+ \lambda\sum_{v=1}^V(CX_vX_v^{k\top}-X_vX_v^{k\top}) 
\end{split}
\end{equation}

Analogously, the derivative of $\mathcal{L}$ with respect to $W$ is:

\begin{equation} \label{equ-gW}
gW = \sum_{d=1}^D (H\odot W)F_dCC^{\top}F_d^{\top}-H\odot (F_dC^{\top}F_d^{\top})
\end{equation}

Let $\alpha$ and $T$ be the step-size and number of iterations. In each stage, we adopt the following update rules:

\begin{equation} \label{equ-updateC}
C_{t+1} = C_{t}-\alpha\frac{gC_t}{\left \| gC_t \right \|_F}
\end{equation}

\begin{equation} \label{equ-updateW}
W_{t+1} = W_{t}-\alpha\frac{gW_t}{\left \| gW_t \right \|_F}
\end{equation}

\begin{equation} \label{equ-updateF}
\begin{split}
F_{d(t+1)} = Y\odot F_{d(t)} + (\bm 1 - Y)\odot ((H\odot W_t)F_{d(t)}C_t)
\end{split}
\end{equation}
where $t = 1, 2, \cdots, T$.

Based on the above update equations, the iterative learning and
prediction process for MLC-PPF are summarized in Algorithm 1.

\begin{algorithm}[]
	\caption{MLC-PPF}
	\LinesNumbered 
	\KwIn{PPF matrices [$F_{1},\cdots,X_{D}$]; Mutiple views of statistical data [$X_{1},\cdots,X_{V}$].}
	\KwOut{Prediction $F_D$}
	Initialize $C$: $C \leftarrow (Y\odot (H \odot W)F_D))^ {\dagger}(Y \odot F_D)$ by solving Eq. (\ref{equ-k-nearest}), where $\dagger$ the pseudo-inverse of matrix;\\ Initialize $W$: $W \leftarrow S$, where $S$ is built by Eq. (\ref{equ-similarity})\\
	Construct $H$ by the real geographic location and regional similarity.\\
	\For{\textit{t} = 1 to \textit{T}}{
		\eIf{ $\left | \mathcal{L}_t - \mathcal{L}_{t+1} \right |$ / $\mathcal{L}_t$ $\geq$ $\varepsilon $}{
			
				update $C_{t+1}$ \textbf{By} Equation \ref{equ-updateC}\\
				update $W_{t+1}$ \textbf{By} Equation \ref{equ-updateW}\\	
				update $F_{t+1}$ \textbf{By} Equation \ref{equ-updateF}\\	
						
		}{
			Break
		}
	}
	Return $\hat{F_{D}}$ \textbf{By} Equation \ref{equ-predict}.
\end{algorithm}

\section{Experiments}

\subsection{Data Description}

\begin{itemize}
	\item We describe the transactional dataset used in this paper, which is a large-scale, real-world dataset provided by NSW Sydney Transport. After data cleaning\footnote{We removed the recording errors and UNKNOWN trips, etc.}, the dataset contains above 35 million transactional records covering 194 stations including the city train and ferry stations between 7 Nov 2016 and 11 Dec 2016. We pick the data between 7 Nov. 2016 and 20 Nov. 2016 as the
	training and validation sets (used to tune parameters); the remaining data are used as the test set.
	
	\item The urban statistical data are collected from Australian Bureau of Statistics-2016 (ABS) with four views; those are Economy, Family, Income, and Population. The numbers of the di-mension of four views are 43, 44, 50, 97, respectively.
	
	\item All the transactional dataset across the transport network are mapped into 117 areas to build the flow matrices $F_d$, $d = 1,..., D$. The designation of areas is based on the Australian Statistical Geography Standard for the best practical value. 
\end{itemize}

\subsection{Methods and Metrics}

We use the following five baselines which can learn the flow data by the cross-domain knowledge guidance. Among them, CDTF and WITF are two tensor-factorization-based (TF) methods that can solve the cold-start problem. For NMF, we concatenate the flow matrix with the statistical data. All parameters used in baselines and our method are picked by a grid search approach.

\begin{itemize}
	
	\item \textbf{NMF}: Predict the PPF by the non-negative matrix factorization, which concatenates the flow matrix and the statistical data \cite{lee2001algorithms}.
	
	\item \textbf{MVL-IV}: A state-of-the-art multi-view learning method
	based on the matrix co-factorization, it learns the same coefficient
	matrix to connect multiple views \cite{xu2015multi}. In this method, we set the flow matrix as one of the views, other views are from the ABS dataset.

	\item \textbf{LS-KNN}: Latent similarity \textit{k}-nearest neighbors. After calculating the latent similarities among areas by Eq. (2), we pick \textit{k}-nearest neighbors of the target areas, and use average crowd flows of these neighbors as an estimate (\textit{k}=4).	
	
	\item \textbf{CDTF}: A state-of-the-art TF method to learn the cross-domain knowledge \cite{hu2013personalized}.  
	
	\item \textbf{WITF}: A weighted irregular TF method which is similar as the CDTF \cite{hu2016learning}. For CDTF and WIFT, we leverage the passenger flow and ABS data to construct the tensor.

\end{itemize}

\begin{table*}[]
	\centering
	\caption{Comparisons with different time periods. We report the average mean absolute errors (MAE) and Normalized Root Mean Square Error (NRMSE) among various methods. The target areas occupied 20\% of the total set. Best results are bold.}
	\label{table1}
	
\begin{tabular}{c|cc|cc|cc|cc}
	\hline
	\multicolumn{1}{c|}{\multirow{2}{*}{Methods}} & \multicolumn{2}{c|}{Morning Rush Hour}              & \multicolumn{2}{c|}{Afternoon Rush Hour}             & \multicolumn{2}{c|}{Non-rush Hour}                  & \multicolumn{2}{c}{Average}    \\
	\multicolumn{1}{c|}{}                         & ~~\textbf{MAE}~~  & \multicolumn{1}{c|}{\textbf{NRMSE}} & ~~\textbf{MAE} ~~  & \multicolumn{1}{c|}{\textbf{NRMSE}} & ~~\textbf{MAE} ~~ & \multicolumn{1}{c|}{\textbf{NRMSE}} & ~~\textbf{MAE}~~  & \textbf{NRMSE}  \\ \hline
	NMF                                           & 124.50        & 30.78\%                             & 117.92         & 37.13\%                             & 89.11         & 28.44\%                             & 110.51        & 32.12\%         \\
	MVL-IV                                        & 108.31        & 29.50\%                             & 101.55         & 29.78\%                             & 92.05         & 27.54\%                             & 100.64        & 28.94\%         \\
	CDTF                                          & 75.15         & 22.43\%                             & 84.02          & 25.93\%                             & 67.78         & 19.37\%                             & 75.65         & 22.58\%         \\
	WITF                                          & 69.30         & 18.73\%                             & 72.06          & 19.45\%                             & 62.57         & 17.26\%                             & 67.98         & 18.48\%         \\
	LS-KNN                                         & 19.89         & 5.42\%                              & 20.20          & 7.67\%                              & 23.51       & 7.94\%                              & 21.20         & 7.01\%          \\ \hline
	MLC-PPF                                       & \textbf{9.84} & \textbf{2.30\%}                     & \textbf{11.47} & \textbf{3.12\%}                     & \textbf{8.22} & \textbf{1.21\%}                     & \textbf{9.84} & \textbf{2.21\%} \\ \hline
\end{tabular}
\end{table*}

\begin{table*}[]
	
	\centering
	\caption{Comparisons with different removing ratios. We report MAE and NRMSE through all test data.}
	\label{table2}
\begin{tabular}{c|cc|cc|cc|cc}
	\hline
	\multirow{2}{*}{Methods} & \multicolumn{2}{c|}{5\%}        & \multicolumn{2}{c|}{10\%}       & \multicolumn{2}{c|}{15\%}       & \multicolumn{2}{c}{25\%}        \\
	& ~~\textbf{MAE}~~  & \textbf{NRMSE}  & ~~\textbf{MAE} ~~ & \textbf{NRMSE}  & ~~\textbf{MAE} ~~ & \textbf{NRMSE}  & ~~\textbf{MAE}~~   & \textbf{NRMSE}  \\ \hline
	NMF                      & 88.11         & 20.70\%         & 90.23         & 21.66\%         & 107.44        & 26.52\%         & 128.70         & 30.11\%         \\
	MVL-IV                   & 92.75         & 20.36\%         & 90.40         & 20.75\%         & 99.01         & 23.79\%         & 101.93         & 32.40\%         \\
	CDTF                     & 59.25         & 18.90\%         & 61.25         & 19.26\%         & 68.07         & 22.00\%         & 79.06          & 26.35\%         \\
	WITF                     & 60.41         & 18.23\%         & 60.72         & 19.77\%         & 61.18         & 19.23\%         & 71.07          & 20.73\%         \\
	LS-KNN                    & 13.69         & 4.72\%          & 16.45         & 5.01\%          & 18.51         & 5.44\%          & 23.99          & 9.25\%          \\ \hline
	MLC-PPF                  & \textbf{8.64} & \textbf{1.37\%} & \textbf{8.80} & \textbf{1.20\%} & \textbf{9.73} & \textbf{1.90\%} & \textbf{11.07} & \textbf{2.34\%} \\ \hline
\end{tabular}
\end{table*}

\textbf{Metrics.} We used the two most widely used evaluation metric to measure the
PPF prediction quality. They are \textit{Mean Absolute Error} (MAE) and \textit{Normalized Root Mean Square Error} (NRMSE).

	\begin{displaymath}
	\begin{split}
	&MAE = \frac{\sum_{i,j=1}^{M}|f_{ij}-\hat{f}_{ij}|}{M} \\NRMSE &= \frac{100\%}{nval}\sqrt{\frac{1}{M}\sum_{i,j=1}^{M}(f_{ij}-\hat{f}_{ij})^2}
	\end{split}
	\end{displaymath}

where $\hat{f}_{ij}$ is a forecasting passenger flow from $i$-th area to $j$-th; and $f_{ij}$ is the ground truth; $M$ is the number of predictions; $nval=max(f_{ij})-min(f_{ij})$.

\subsection{Comparisons on Different Time Periods}

The first set of experiments is designed to assess the performance on different time periods. We randomly removed 20\% areas as the target set, and the remaining 80\% areas as the known set. The learning step $\alpha$ is fitted to $10^{-2}$ and there are only two hyper-parameters used in our method, where $k$ and $\lambda$ are chosen from
\{1,2,3,...,10\} and \{$10^{-5},10^{-4}, ...,10^{5}$\} respectively. We repeat the experiment 20 times with random initialization and report the average results.

Experimental results are presented in Table \ref{table1}. Compared with other approaches, our method achieved the best prediction accuracies on both three time periods. None of the multi-view and cross-domain methods work well because it is hard to capture the relationships between statistical data and the passenger flows. The approach LS-KNN performs better than other baselines, which illustrates that the PPF prediction problem has a strong spatial correlation property. In summary, the proposed method is a well-designed model for PPF prediction, which outperforms the other available baselines because it considers the localized correlations and the cross-domain knowledge simultaneously.

\subsection{Comparisons on Various Missing Ratios}

In this experiment, we evaluate how performance will change with
varied number of target areas. We randomly pick 5\%, 10\%, 15\%, 25\% areas as the target areas, and run 20 times to report the average errors. The test period is in the morning rush hours. The performances of different methods are summarized in Table \ref{table2}. 

It is apparent that the experimental results lead to
similar conclusions to the first comparison. Our model, MLC-PPF, significantly outperforms all other comparative methods over all testing sets. The performances of MLC-PPF in 5\% dataset are very close to that of in 10\%, which illustrates that the 90\% remaining area set can learn a satisfied localized correlation and make accurate PPF predictions. In the real-world application, the proportion of target areas is usually small since only a few areas are suitable for constructing a new station.

\begin{figure}
	\centering
	\subfloat[Factor $k$]{\includegraphics[width=0.5\linewidth]{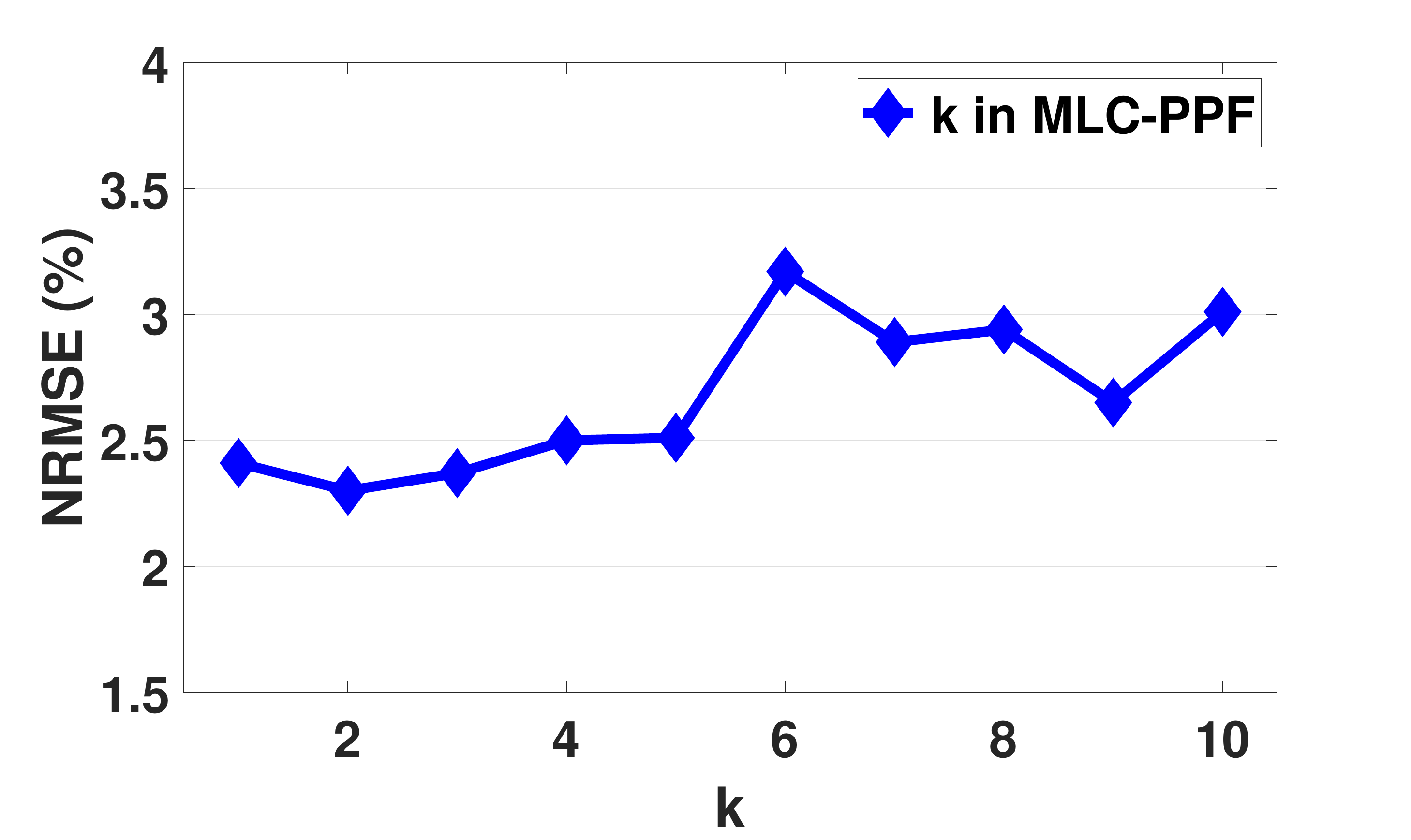}} 
	\subfloat[Factor $\lambda$]{\includegraphics[width=0.5\linewidth]{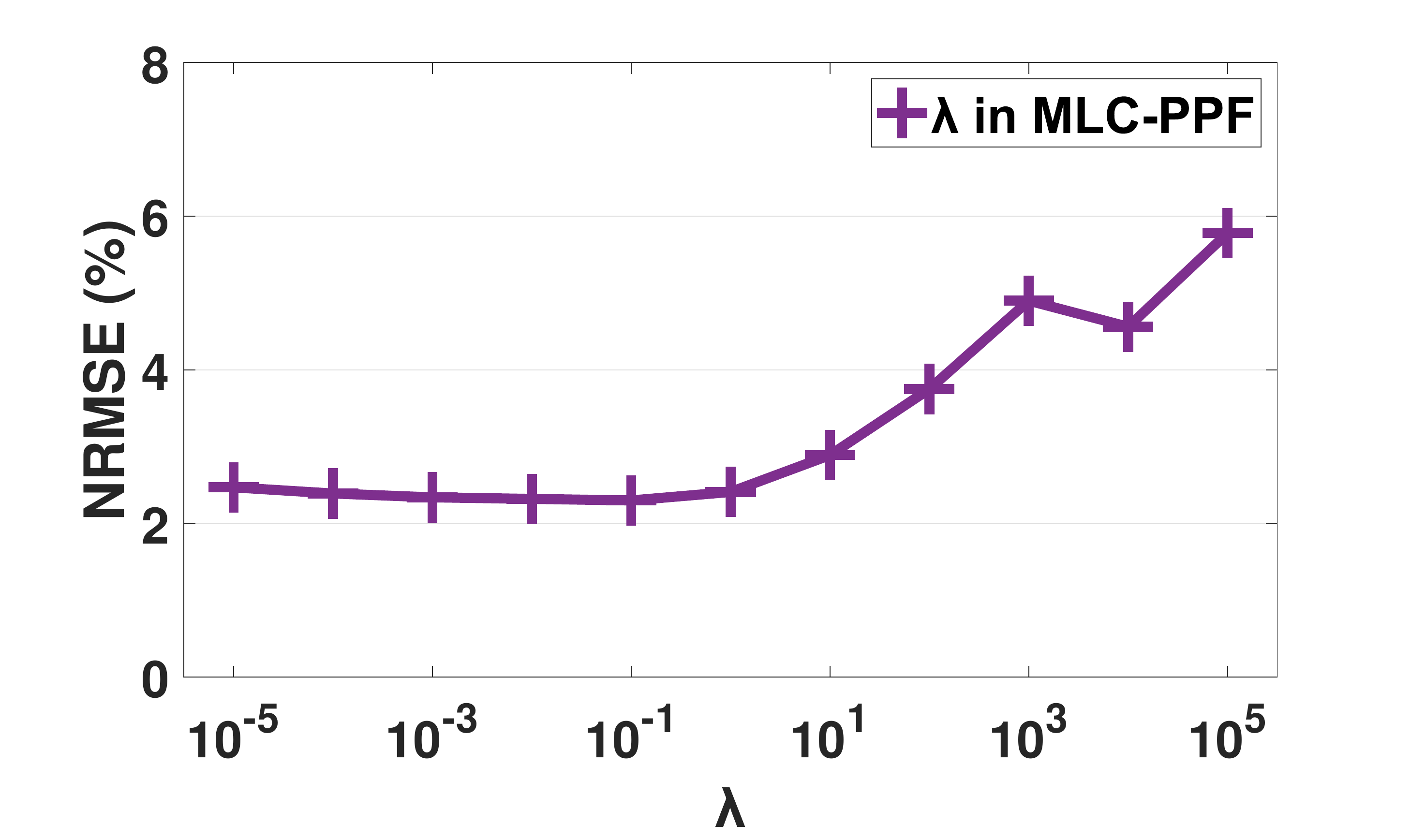}}
	\caption{Effect of Parameters} 
	\label{fig-parameters}
\end{figure}

\begin{figure*}[t]
	\centering
	\includegraphics[width=0.9\linewidth]{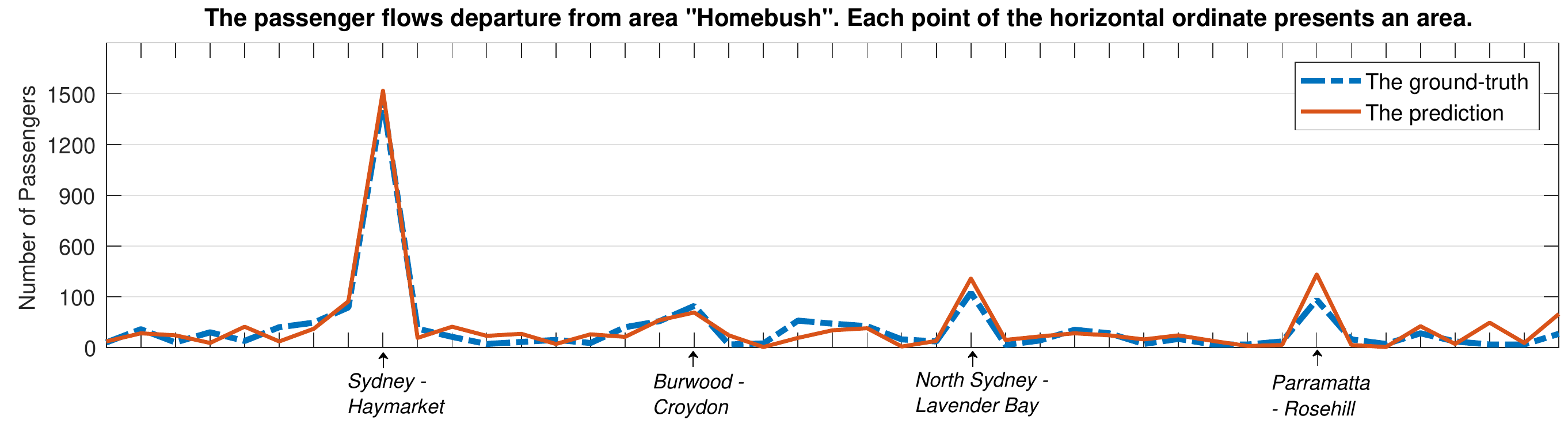}
	\caption{The case study. This figure shows the passenger flow prediction that departure from ``Homebush'' to other areas. To keep figure
		clear, we only draw our method and the ground-truth because
		other baselines perform far worse than the MLC-PPF.}	
	\label{figure_case}
\end{figure*}

\subsection{Parameter Analysis}

In this section, we analyze the effects of two hyper-parameters used in this paper, where $k$ is the number of nearest neighborhoods, and the regularization factor $\lambda$ controls the strength of guidance from $ABS$ data. 

Figure \ref{fig-parameters}(a) shows the different performances with a varying setting for $k$. 
For each area, the correlation matrix $C$ only learns the transform from these neighborhoods. As can be shown in the results,  $k$ = 2 is the best choice for our method, and the model is not sensitive to $k$ between 1 and 5.

Figure \ref{fig-parameters}(b) represents various results by varying $\lambda$. 
$\lambda$=$10^{-1}$ achieves the best results, and the performances are stable when choosing between [$10^{-5}$, $10^0$]. 

In a summary, the parameters used in this paper are benefit
to the improvement of our models. MLC-PPF is stable because it is insensitive to parameters. 

\subsection{Case Study}
We display a PPF prediction result of one target area in this section. In this case, the area ``Homebush'' is treated as the target area. For better visualization, we only remain the areas where the number of arrived passengers is greater than 5. 

As shown in Figure \ref{figure_case}, our model yields a great prediction result compared with the ground-truth, especially in some main areas of Sydney, such as the central area ``Sydney-Haymarket'', ``Burwood-Croydon'', ``North Sydney-Lavender Bay'' and ``Parramatta-Rosehill''. The case study demonstrates the effectiveness of our method for the PPF prediction.

\subsection{Transfer to the Cold-start Problem}

As we have emphasized, our strategy can provide a new perspective to address the classic cold-start problem in the recommender system. This set of experiments is designed to assess the transferability of our model. 

We choose a very famous dataset from Amazon to do the evaluation, in which the dataset contains 1,555,170 users and 1-5 scaled ratings over 548,552 different products covering four
domains: books, music CDs, DVDs and videos \cite{hu2013personalized}. We randomly remove the 20\% users from target domains to simulate the cold-start situation. Three baselines are used in this comparison. CMF is an effective method based on the collective matrix factorization which
couples rating matrices for all domains on the User dimension \cite{singh2008relational}. CDTF and WITF are two tensor-factorization-based cross-domain recommendation methods, they devise a strategy to transform original data into a cubical tensor that can better capture the interactions between user factors and item factors \cite{hu2013personalized,hu2016learning}. In this experiment, we leverage the information excluding target domain to build the $k$-nearest indicator matrix $H$.

Table \ref{table_cold-start} shows the results of our methods together with some state-of-the-art approaches. MLC-PPF can achieve the greatest accuracies for the target domains, which illustrates that our method is able to solve the unacquainted world phenomenon and give inspiration for relevant tasks. Despite the effectiveness of our methods, we should admit that there is a limitation of MLC-PPF. MLC-PPF only can make the prediction when its $k$-neighbors have ratings. However, based on the test results, the predicted ratings are reliable and able to make the recommendation.

\begin{table}[]
	
	\centering
	\small
	\caption{Transfer to the cold-start problem. We report the MAE of all test methods.}
	\label{table_cold-start}
	\begin{tabular}{c|cccc}
		\hline
		Target Domain & CMF & CDTF & WITF & MLC-PPF \\ \hline
		Book          &  0.834   &  0.755    &  0.740    &  \textbf{0.396 }      \\
		Music         &  0.847   &  0.779    &  0.716    &  \textbf{0.582 }       \\ \hline
	\end{tabular}
\end{table}

\section{Conclusion}

In this paper, we proposed an effective method for the potential passenger flow prediction, which is a novel study that brings benefits to the urban transportation development. To address this spatio-temporal problem, we design a multi-view localized correlation learning model (MLC-PPF) for the PPF prediction. The $k$-nearest indicator matrix $H$ is constructed by the real geographical neighbors and regional properties. MLC-PPF can learn the correlations between each known area and its \textit{k}-nearest neighbors with the cross-domain knowledge guidance. We evaluate the performance of our method with a set of well-designed experiments. All empirical results not only demonstrate that the proposed model outperforms all the other methods in the PPF prediction task, but also represent the capability of tackling the cold-start problem in recommender system.

\bibliographystyle{aaai}
\bibliography{aaai20}

\end{document}